\documentclass[letterpaper]{article} 
\usepackage[submission]{aaai23}  
\usepackage{times}  
\usepackage{helvet}  
\usepackage{courier}  
\usepackage[hyphens]{url}  
\usepackage{graphicx} 
\usepackage{natbib}  
\usepackage{caption} 
\usepackage{algorithm}
\usepackage{algorithmic}
\usepackage{amsmath}
\usepackage{appendix}
\usepackage{booktabs}
\usepackage{adjustbox}
\usepackage{amssymb}

\usepackage{newfloat}
\usepackage{listings}
\DeclareCaptionStyle{ruled}{labelfont=normalfont,labelsep=colon,strut=off} 
\floatstyle{ruled}
\newfloat{listing}{tb}{lst}{}
\floatname{listing}{Listing}
\title{Domain Generalization via Contrastive Causal Learning}
\author {
    Qiaowei Miao,\textsuperscript{\rm 1}
    Junkun Yuan, \textsuperscript{\rm 1}
    Kun Kuang \textsuperscript{\rm 1}
}
\affiliations {
    \textsuperscript{\rm 1} Zhejiang University\\
    qiaoweimiao@zju.edu.cn, yuanjk@zju.edu.cn, kunkuang@zju.edu.cn
}
\usepackage{bibentry}

\begin{document}

\maketitle

\frenchspacing

\begin{abstract}
Domain Generalization (DG) aims to learn a model that can generalize well to unseen target domains from a set of source domains. With the idea of invariant causal mechanism, a lot of efforts have been put into learning robust causal effects which are determined by the object yet insensitive to the domain changes. Despite the invariance of causal effects, they are difficult to be quantified and optimized.
Inspired by the ability that humans adapt to new environments by prior knowledge, We develop a novel Contrastive Causal Model (CCM) to transfer unseen images to taught knowledge which are the features of seen images, and quantify the causal effects based on taught knowledge. Considering the transfer is affected by domain shifts in DG, we propose a more inclusive causal graph to describe DG task. Based on this causal graph, CCM controls the domain factor to cut off excess causal paths and uses the remaining part to calculate the causal effects of images to labels via the front-door criterion.
Specifically, CCM is composed of three components: \textit{(i)} domain-conditioned supervised learning which teaches CCM the correlation between images and labels, \textit{(ii)}  causal effect learning which helps CCM measure the true causal effects of images to labels, \textit{(iii)} contrastive similarity learning which clusters the features of images that belong to the same class and provides the quantification of similarity. Finally, we test the performance of CCM on multiple datasets including \textit{PACS}, \textit{OfficeHome}, and \textit{TerraIncognita}. The extensive experiments demonstrate that CCM surpasses the previous DG methods with clear margins.
 
\end{abstract}

\section{Introduction}

Humans have the ability to solve specific problems with the help of previous knowledge. This generalization capability helps humans take advantage of stable causal effects to adapt to the environment shift. While deep learning has achieved great success in a wide range of real\mbox{-}world applications, due to the lack of out\mbox{-}of\mbox{-}distribution (OOD) generalization ability \cite{krueger2021out,sun2020test,zhang2021deep}, it suffers from a catastrophic performance degradation problem, especially when deployed in new environments with changing distributions. Although Domain Adaptation (DA) algorithms \cite{fu2021transferable, li2021learning, li2021transferable, li2021semantic} support models to adapt to various target domains, different target domains need corresponding domain adaptation processes. In order to deal with domain shift problems, Domain Generalization (DG) \cite{zhang2021deep, chen2021style, liu2021learning, sun2021recovering, mahajan2021domain,wald2021calibration} is introduced, which aims to learn stable knowledge from multiple source domains and train a generalizable model directly to unseen target domains. 

Increasing works on DG have been proposed with a variety of strategies like data augmentation \cite{Carlucci2019DomainGB,Wang2020LearningFE, DBLP:conf/aaai/ZhouYHX20,zhou2020learning,zhou2021domain},
meta-learning \cite{balaji2018metareg,li2018learning,dou2019domain, Li2019EpisodicTF,li2019feature}, invariant representation learning \cite{Zhao2020DomainGV, Matsuura2020DomainGU, li2018deep, DBLP:conf/aaai/LiGTLT18}, et al. While promising performance has been achieved by these methods, they might try to model the statistical dependence between the input features and the labels, hence could be biased by the spurious correlation in data \cite{liu2021learning}. With the idea of invariant causal mechanism, increasing attention has been paid to causality-inspired generalization learning \cite{wald2021calibration, liu2021learning, sun2021recovering, mahajan2021domain}. For example, Liu et al. \cite{liu2021learning} introduce their causal semantic generative model to learn semantic factors and variation factors separately via variational Bayesian. Sun et al. \cite{sun2021recovering} further introduce a domain variable and an unobserved confounder to describe a latent causal invariant model. Mahajan et al. \cite{mahajan2021domain} consider high-level causal features and domain-dependent features while the labels are only determined by the former. Each of the casual graphs given by them has a different focus, but there are commonalities among them.

\begin{figure}[t]
\centering
\includegraphics[width=1\linewidth]{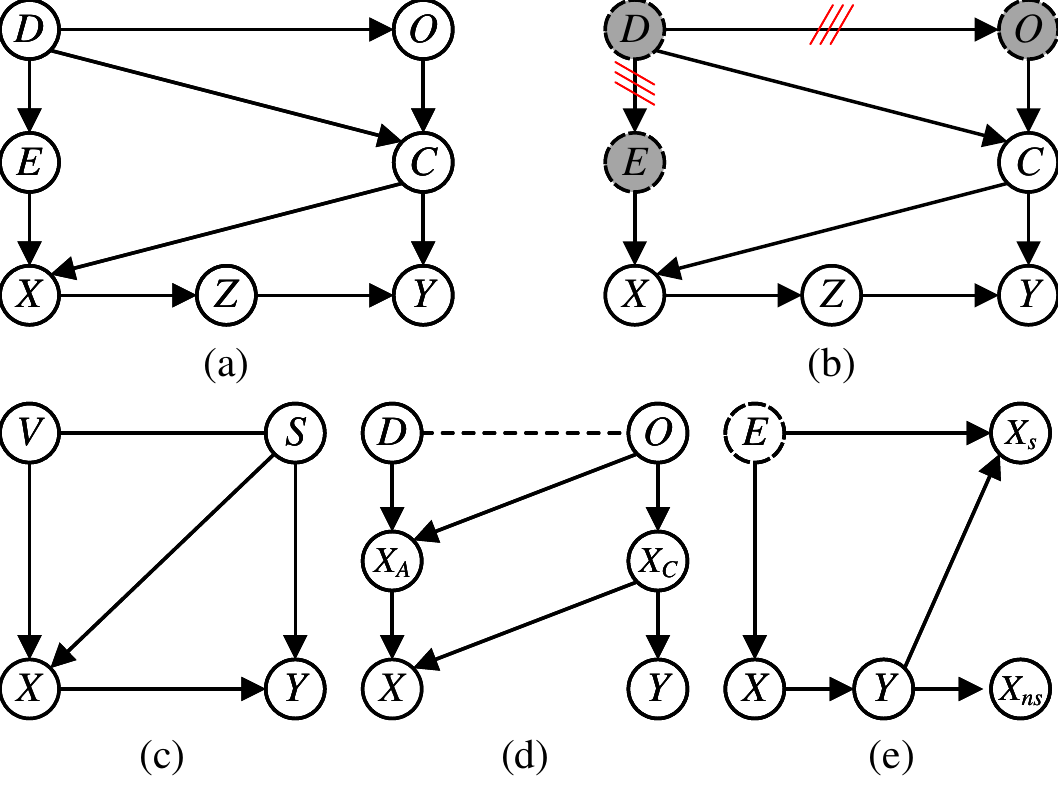}
\caption{Comparisons between the causal graphs of CCM (a), (b) and the previous methods (c) \cite{liu2021learning, sun2021recovering}, (d) \cite{mahajan2021domain}, and (e) \cite{wald2021calibration}. In Figure (a), based on Figure (c), Figure (d), and Figure (e), we add prior knowledge $Z$ as a bridge to link unseen image $X$ and label $Y$ and make domain $D$ point to object $O$ and category factor $C$ to explain the limitations of source domain on models in DG. In Figure (b), by controlling domain $D$, the remaining part is a standard causal graph that can calculate causal effects from $X$ to $Y$ via the front-door criterion.
}



\label{fig:causal-graph}
\end{figure}

In this paper, we propose a causal graph to formalize the DG problem from a novel perspective as shown in Figure \ref{fig:causal-graph} (a) and Figure \ref{fig:causal-graph} (b). To create a data sample in a certain environment, e.g., an image of a polar bear in the Arctic and the corresponding label of the bear, We can guess from the domain $D$ of the image that the object $O$ is a hardy animal (i.e. $D \rightarrow O$). The hardy animal that lives in the Arctic may be the polar bear (i.e. $D \rightarrow C \leftarrow O$). The domain factor not only affects object $O$ and category $C$, but also provides background features $E$ to the image $X$ (i.e. $D \rightarrow E \rightarrow X$). Combining the dual information of background $E$ and category $C$, the image $X$ is captured and transformed into prior knowledge (e.g. the seen images of brown bears) to predict its label $Y$ (i.e. $X \rightarrow Z \rightarrow Y$). And the label $Y$ is determined based on the match between knowledge $Z$ and category $C$. 
Compared to Figure \ref{fig:causal-graph} (c), Figure \ref{fig:causal-graph} (d) and Figure \ref{fig:causal-graph} (e), we add prior knowledge $Z$ as a bridge to link unseen image $X$ and label $Y$. The the relationships between domain $D$ and other factors can be explained more clearly by figure \ref{fig:causal-graph} (b). In Figure \ref{fig:causal-graph} (a), the domain $D$ as a confounder disturbs models to learn causal effects from image $X$ to labels $Y$. So we control domain $D$ to cut off $D \rightarrow O$ and $D \rightarrow E$, and the remaining part is a standard causal graph that can calculate causal effects from $X$ to $Y$ via the front-door criterion, as shown in Figure \ref{fig:causal-graph} (b).

To learn the causal effects of $X$ to $Y$ shown in Figure \ref{fig:causal-graph} (b), we introduce the front-door criterion. It splits the causal effects of $P(Y|do(X))$ into the estimation of three parts: $P(X)$, $P(Z|X)$, and $P(Y|Z,X)$. Furthermore, to permit stable distribution estimation under causal learning, we further design a contrastive training paradigm that calibrates the learning process with the similarity of the current and previous knowledge to strengthen true causal effects. 

Our main contributions are summarized as follows. 
(i) We develop a Contrastive Causal Model to transfer unseen images into taught knowledge that, and quantify the causal effects between images and labels based on taught knowledge.
(ii) We propose  an inclusive causal graph that can explain the inference of domain in the DG task. Based on this graph, our model cuts off the excess causal paths and quantifies the causal effects between images and labels via the front-door criterion.
(iii) Extensive experiments on public benchmark datasets demonstrate the effectiveness and superiority of our method. 

\section{Related Work}
\subsection{Domain Generalization}
Domain generalization (DG) aims to learn from multiple source domains a model that can perform well on unseen target domains. Data augmentation-based methods \cite{volpi2018generalizing, shankar2018generalizing, Carlucci2019DomainGB, Wang2020LearningFE, DBLP:conf/aaai/ZhouYHX20, zhou2020learning, zhou2021domain} try to improve the generalization robustness of the model by learning from the data with novel distributions. Among them, some work \cite{volpi2018generalizing, shankar2018generalizing} generates new data based on model gradient and leverages it to train a model for boosting its robustness. While others \cite{Wang2020LearningFE, Carlucci2019DomainGB} introduce an interesting jigsaw puzzle strategy that improves model out-of-distribution generalization via self-supervised learning. Adversarial training \cite{DBLP:conf/aaai/ZhouYHX20, zhou2020learning} is also employed to generate  data with various styles yet consistent semantic information. Meta-learning \cite{balaji2018metareg, li2018learning, dou2019domain, Li2019EpisodicTF, li2019feature} is also a popular topic in DG. The idea is similar to the problem setting of DG: learning from the known and preparing for inference from the unknown. However, it might not be easy to design effective meta-learning strategies for training a generalizable model. Another conventional direction is to perform invariant representation learning \cite{Zhao2020DomainGV, Matsuura2020DomainGU, li2018deep, DBLP:conf/aaai/LiGTLT18}. These methods try to learn the feature representations that are discriminative for the classification task but invariant to the domain changes. For example, \cite{Zhao2020DomainGV} proposes conditional entropy regularization to extract effective conditional invariant feature representations. While favorable results have been achieved by these approaches, they might try to model the statistical dependence between the input features and the labels, hence could be biased by the spurious correlation \cite{liu2021learning}. 

\subsection{Domain Generalization with Causality}
In this paper, we assume the data is generated from the root factors of the object $O$ and domain $D$ as shown in Figure \ref{fig:causal-graph} (a). The class features $C$ control both the input feature $X$ and the label $Y$, meanwhile, the environment feature $E$ only affects $X$. We aim to learn an informative representation from $X$ to predict $Y$. 
\cite{liu2021learning} proposes a causal semantic generative model (see Figure \ref{fig:causal-graph} (c)). It separates the latent semantic factor $S$ and variation factor $V$ from data, where only the former causes the change in label $Y$. Similarly, \cite{sun2021recovering} introduces latent causal invariant models based on the same causal model structure. Their semantic factor $S$ and variation factor $V$ are similar to the class feature $C$ and the environment feature $E$ in our causal graph respectively, while we further show their causal relationship with domain and object. 
\cite{mahajan2021domain} proposes a causal graph with the domain $D$ and object $O$ which is similar to ours, as shown in Figure \ref{fig:causal-graph} (d). It assumes that the input feature $X$ is determined by causal feature $X_{C}$ and domain-dependent feature $X_{A}$, and the label $Y$ is determined by $X_{C}$. Actually, the representation $Z$ (Figure \ref{fig:causal-graph} (a)) that we aim to learn is to capture the information of causal feature $X_{C}$ (Figure \ref{fig:causal-graph} (d)). 
\cite{wald2021calibration} puts forward model calibration for the source domains based on the graph in Figure \ref{fig:causal-graph} (e). It splits the information in the label $Y$ into anti-causal spurious features $X_{s}$ and anti-causal non-spurious features $X_{ns}$, where the information of the latter is aimed to learn as the representation $Z$ in our causal graph. Therefore, our proposed causal graph can be seen as a uniform structure of the previous methods and is generally compatible with the structure of the previous works. 

\subsection{Contrastive Learning}
Recently, contrastive learning \cite{wang2021understanding, chen2021exploring, DBLP:conf/icml/ChenK0H20, he2020momentum} as an unsupervised learning paradigm has drawn increasing attention due to its excellent representation learning ability. 
The goal of contrastive learning is to gather similar samples and diverse samples that are far from each other.
For example, \cite{he2020momentum} introduces the popular MoCo framework, which builds dynamic dictionaries and learns the representations based on a contrastive loss. 

\begin{figure*}[t]
\centering
\includegraphics[width=1\textwidth]{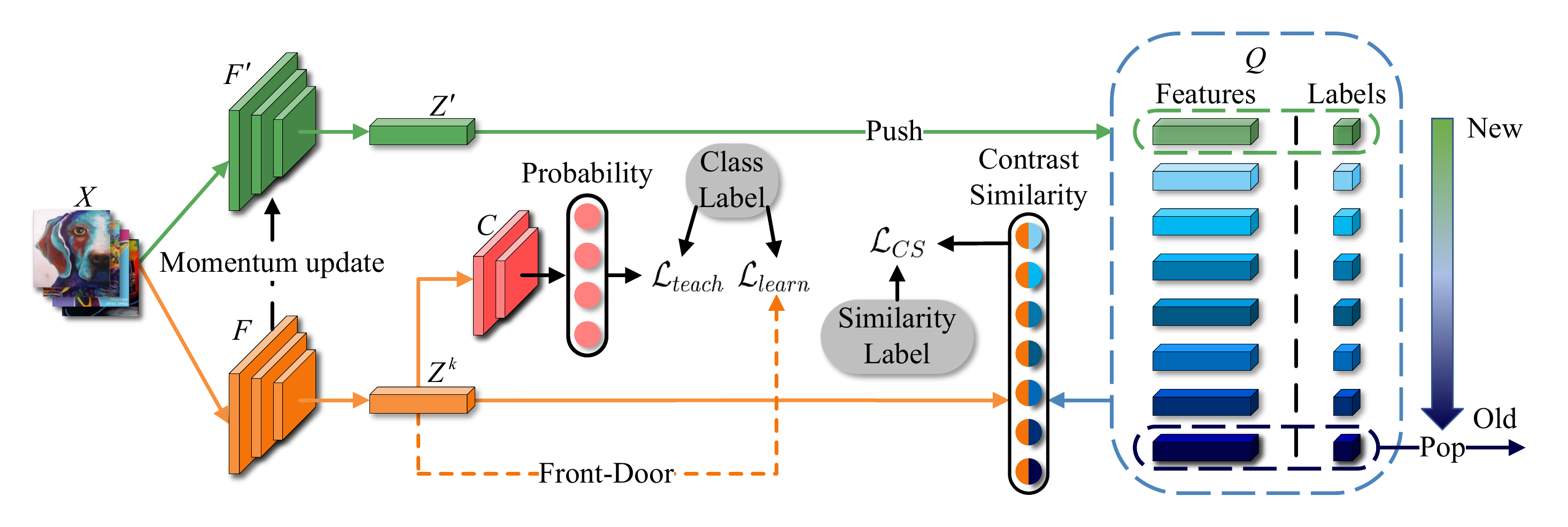}
\caption{The framework of CCM. 
The teacher $F$ and student $F'$ have the same structure, and the student $F'$ gets experiences from teacher $F$ by momentum update. The teacher $F$ extract the features $Z^{k}$ of images $X^{k}$ and send them to classifier $C$ to calculate $\mathcal{L}_{teach}$. The features $Z^{k}$ are also used to calculate $\mathcal{L}_{learn}$ via the front-door criterion with knowledge queue $Q$. To quantify the causal effect more accurately, we measure the contrastive similarity between features $Z^{k}$ and the features in the knowledge queue $Q$. And optimizing $\mathcal{L}_{CS}$ can cluster the features with the same category. Finally, the student $F'$ obtain historical features $Z'$ and pushes them into the knowledge queue $Q$. The oldest features and their labels in the knowledge queue $Q$ are popped.
}
\label{figs:CCM}
\end{figure*}

\section{Proposed Method}
To introduce CCM more clearly, we give some necessary notations first. We donate the joint space of the input image and the label as $\mathcal{X} \times \mathcal{Y}$. There are $k$ source domains with different statistical distributions defined on the joint space. The $k$th source domain is $D_{k} = \{(x_{i}^{k},y_{i}^{k})\}_{i=1}^{N_{k}}$. The teacher $F$ and student $F'$ are two backbones with the same structure. The features of images are defined as $Z$. CCM has a classifier $C$ and a knowledge queue $Q$, which stores features and labels.

The Contrastive Causal Model (CCM) addresses the domain generalization problem by introducing contrastive similarity to convert new images to previous knowledge and increase the percentage of causal effects from images to labels with the help of the front-door criterion. In the following, we introduce the structural causal model of CCM first. And then, we introduce the structure of CCM specifically from three main parts:  domain-conditioned supervised learning, causal effect learning, and contrastive similarity learning.
\subsection{Structural Causal Model of CCM}

There is a series of prior works \cite{christiansen2021causal,yuan2021learning,mahajan2021domain, chen2021style,sun2021recovering, li2021invariant} investigate domain generalization task and provide their own approaches in causal view. While these methods strengthen the causal effects between images to labels, the spurious correlation given by domain does not get enough attention. Using the structural causal model (SCM) can explain this weakness obviously. In \cite{christiansen2021causal,yuan2021learning}, their SCMs show the direct causal path from image $X$ to label $Y$. However, \cite{christiansen2021causal} ignores the intermediate product $Z$ of the feature extractor and classifier. \cite{yuan2021learning} regard $Z$ as domain-invariant relationship. In our SCM, the causal path $X \rightarrow Z \rightarrow  Y$ implies extracting true causal features $Z$ from image $X$ by feature extractor $F$ and predicting the label $Y$ based on $Z$ by classifier $C$. In addition, several recent works\cite{mahajan2021domain, chen2021style,sun2021recovering, li2021invariant} have further identified other hidden causal relationship. These works introduce domain variable $D$. $D$ determines the domain identity $E$ which independent of category information, representing in Fig \ref{fig:causal-graph}(a) as $D \rightarrow E$. Especially in works\cite{li2021invariant, mahajan2021domain, chen2021style}, these SCMs introduce variables that are only relevant to objects. Like the path from object $O$ to category factor $C$. By mixing the object factor and domain factor, the image $X$ is generated by causal path $E \rightarrow X \leftarrow C$. But in a fixed domain, $D$ also influence $O$ and $C$. For example, we can speculate about the presence of the bear object based on the information that the location is the North Pole. But the location cannot be inferred to be the North Pole based solely on the presence of the bear object. So we add the path $D \leftarrow O$. Along these lines, the Arctic environment and the presence of bear targets led to a more refined category factor of polar bears rather than black bears. So it has a path from $D$ to $C$. The SCMs proposed by previous works are organized as Fig \ref{fig:causal-graph}(a). 

In Fig \ref{fig:causal-graph} (a), there four causal effect paths from image $X$ to label $Y$: 
(\romannumeral1) $X \rightarrow Z \rightarrow Y$, 
(\romannumeral2) $X \leftarrow C \rightarrow Y$, 
(\romannumeral3) $X \leftarrow E \leftarrow D \rightarrow C \rightarrow Y$, 
(\romannumeral4) $X \leftarrow E \leftarrow D \rightarrow O \rightarrow C \rightarrow Y$.
Path $X \rightarrow Z \rightarrow Y$ is the main causal effect path. If other paths do not exist, using path $X \rightarrow Z \rightarrow Y$ can learn the pure causal relationship between $X$ to $Y$. So how cutting off the excess path is important. In Fig \ref{fig:causal-graph} (b), we control domain factor $D$  to cut off $D \rightarrow E$ and $D \rightarrow O$ (i.e. cut off $X \leftarrow E \leftarrow D \rightarrow C \rightarrow Y$ and $X \leftarrow E \leftarrow D \rightarrow O \rightarrow C \rightarrow Y$). This operation is equivalent to slicing the source domain data for different sub-domains. Furthermore, with the change of images, only the category factor $O$ disturbs models to learn causal effects of $X \rightarrow Z \rightarrow Y$ in the fixed domain. 
Due to the severance of the causal paths, the category factor $C$ is a confounder of $X$ and $Y$ on the remaining causal graph and is suitable to use the front-door criterion to remove its interference.

\subsection{Domain-Conditioned Supervised Learning}
Conventional supervised learning directly minimizes the empirical risk of the training data to learn the relationship between the features of images and labels. Here, we fix the domain factor to cut off the confounding effect of the domain as shown in Figure \ref{fig:causal-graph} (b). Thus, it turns out to be domain-conditioned supervised learning. And we minimize the cross-entropy loss conditioned on the domain for supervised learning. That is,
\begin{equation}
    \mathcal{L}_{teach}^{D_{k}}=-\frac{1}{N_{k}}\sum_{i=1}^{N_{k}} y_{i}^{D_{k}} \textrm{log} C(F(x_{i}^{D_{j}}))
    \label{teach}
\end{equation}
The teaching $\mathcal{L}_{teach}$ helps CCM learn the correlation of image to label across different domains, which contain causal effects from image $X$ to label $Y$. In other words, casual effects are parts of the correlation. If the model only allows using causal effects for inference without teaching it the correlation between image and label, the task will be too difficult for a model to solve. This point is verified in our ablation experiments.

\subsection{Causal Effect Learning}
After teaching the model about correlation, we desire to measure how much CCM learns the causal effects between the input and the label. Here, we introduce the front-door criterion to measure the causal effects and increase the percentage of them via $do(\cdot)$ operation:

\begin{align}\label{front-door}
 &P(Y|do(X))  \\\notag
=&\sum_{z}P(Z=z|X) \\\notag
 &\sum_{x}[P(X=x)P(Y=y|Z=z,X=x)]
\end{align}

Take CCM as an example, $X$ is the set of images, and $Y$ is the set containing their corresponding labels. Moreover, $Z$ is the prior knowledge learned by the images that the model has been taught. Specifically, the causal path $X \rightarrow Z \rightarrow Y$ means the model converts an unseen image to prior knowledge and predicts its label. This behavior is consistent with the reflection of human adaptation to a new environment.  

There are three main parts in $do(\cdot)$: $P(Z=z|X)$, $P(X=x)$ and $P(Y|Z=z,X=x)$. We will introduce them in order.

In $P(Z=z|X)$, $Z$ is a symbol of prior knowledge that has been taught. $Z$ is also an intermediary on the causal path from $X$ to $Y$. Inspired by recent work\cite{he2020momentum}, we also construct a knowledge queue $Q$ to store trained features as prior knowledge to help CCM translate new images into learned knowledge.  The feature set $Z^{Q}=\{z^{Q}_{1}, z^{Q}_{2}, \cdots, z^{Q}_{m}\}$ saved in $Q$ is a subset in feature space $\mathcal{Z}$ and is used for approximating the true distribution of $Z$ in space $\mathcal{Z}$. Not only the features set $Z^{Q}$ are stored in $Q$, but also its corresponding labels set $Y^{Q}=\{y^{Q}_{1}, y^{Q}_{2}, \cdots, y^{Q}_{m}\}$ is saved in $Q$. The effect of $Y^{Q}$ will be introduced specifically in the next section. After obtaining $X$, we define the contrastive similarity to calculate $P(Z=z|X)$  like Eq. \ref{CS}.

\begin{equation} \label{CS}
    M_{CS}(q,k)=\frac{\textrm{norm}(q) \cdot \textrm{norm}(k)^{T}}{\tau \cdot \sqrt{\textrm{d}}}
\end{equation}
where $\tau$ is a temperature hyper-parameter and $\textrm{d}$ the dimension of $q$ and $k$. $q$ and $k$ are two variables feeling in the model to measure contrastive similarity.
More specifically, in Fig \ref{figs:CCM}, we feed a batch of images $X$ of the domain $k$ into teacher $F$ to get corresponding features $Z^{k}$ first. Inspired by \cite{he2020momentum}, there is a student $F'$ which has the same structure as $F$  for obtaining historical features $Z'$. The historical features $Z'$ produced by student $F'$ maintain consistency with $Z^{Q}$ with the help of momentum update from teacher $F$. 
\begin{equation}
    F' = \alpha F' + (1-\alpha) F
    \label{update}
\end{equation}
where $\alpha$ is the momentum ratio. 
It should be noticed that historical features  $Z'$ will be pushed in $Q$ after calculating loss to ensure CCM only use prior knowledge. Also the oldest feature $z_{m}^{Q}$ will be popped out to complete the update of knowledge queue $Q$. We set $Z^{k}$ as the agent of $X$ and use $Z^{k}$ and $Z^{Q}$ to calculate $P(Z=z|X)$. The $P(Z=z|X)$ is as follows:
\begin{equation}
    P(Z=z_{i}^{Q}|X)=\textrm{norm}(\sum_{j=1}^{N}Softmax[M_{CS}(z_{i}^{Q},z_{j}^{k})])
    \label{pzx}
\end{equation}
In Eq. \ref{pzx}, we use L1 normalization to limit the range of output to get $P(Z=z|X)$.

To get $P(X=x)$, we should iterate all images in the real world, which is too hard to achieve. Therefore, we estimate the true distribution of $x$ in $\mathcal{X}$ by batch. It should be noticed that the images in a batch are chosen randomly, so we can measure $P(X=x)$ as $\frac{1}{N}$ directly, which $N$ is the batch size. However, this solution can not represent the relationship precisely between each image in one batch. For example, if there is an extremely similar image set $X_{sim} = \{x_{1},x_{1},\cdots,x_{n}\}$ in a batch, the probability of selecting any one of $X_{sim}$ should be $\frac{n}{N}$ instead of $\frac{1}{N}$. If the image $x$ is more similar to other images in the batch, it means that if the image $x$ is more common, then the corresponding $P(X=x)$ value is higher
We put a batch of images $X$ into teacher $F$ to get a feature set $F(X)$, and then calculate the contrastive similarity between one feature $F(x)$ and features in the feature set $F(X)$ to get $P(X=x)$.

\begin{equation}
    P(X=x) = Softmax[ \sum_{i=1}^{N}M_{CS}(F(x),F(x_{i}))]
    \label{px}
\end{equation}
During the training process, we can get $P(X=x)$ by Eq. \ref{px}. In the inference phase, although the images are combined into a batch and fed into CCM, each image should be regarded as an individual that can not access any other images in this batch. So in the non-training period we set $P(X=x)=\frac{1}{N}$.

The $P(Y|Z=z,X=x)$  needs to iterate $z$ and $x$ as input to compute. Inspired by \cite{yang2021causal}, we parameterize the predictive $P(Y|Z=z,X=x)$ as a model $G$. However, the $E$ of CCM in the implementation details is different. We regard $F(x)$ as the agent of $x$ and reduce the dimensions of $z$ and $F(x)$ to half of the original by model $H$ in order to align the dimensions of these two variables. Subsequently, both of them will be concatenated together as shown in Fig \ref{figs:CCM} and feed in $E$ to get $P(Y|Z=z,X=x)$ through the following equation:
\begin{equation}
\begin{split}
    &P(Y|Z=z,X=x)\\=&Softmax[G(\textrm{concat}(H(z^{Q}),H(F(x))))]
\end{split}
    \label{pyzx}
\end{equation}

And now we can calculate Eq. \ref{front-door} which represents the true causal effects from $X$ to $Y$ contained in the correlation. So we want to minimize $L_{learn}$ to increase the percentage  of causal effects in the domain $D_{k}$ by the following equation:
\begin{equation}
    \mathcal{L}_{learn}^{D_{k}}=-\frac{1}{N_{k}}\sum_{i=1}^{N_{k}} y_{i}^{D_{k}} \textrm{log} M_{FD}(x_{i}^{D_{k}})
    \label{learn}
\end{equation}
where $M_{FD}(\cdot)$ is the symbol for the front-door adjustment formula.

\subsection{Contrastive Similarity Learning}
In Eq. \ref{pzx} and Eq. \ref{px}, $M_{CS}(\cdot)$ is used several times for measure the similarity of two features. Although humans can distinguish which images contain objects in the same category, it is not straightforward to quantify the similarity of two images in the feature space $\mathcal{Z}$. In CCM, we propose Contrastive similarity learning to help the model obtain the ability to cluster the features of images that have the same class. After sending $X$ in $F'$ to get feature set $Z'$, we use $M_{CS}(\cdot)$ to measure the similarity of $Z'$ and $Z^{Q}$. It should be noticed that the labels $Y^{Q}$ of $Z^{Q}$ are saved in $Q$. So, according to $Y$ and $Y^{Q}$, it can separate whether the features belong to the same class or not and get the binary labels $Y_{self}$. Subsequently, we minimize $L_{CS}$ to pull in the contrastive similarity of the features in the space $Z$.
\begin{equation}
    \mathcal{L}_{CS} = -\frac{1}{N_{pair}}\sum_{i=1}^{N_{pair}}\textrm{log} M_{CS}({F}'(x), z^{Q+}_{i})
    \label{l_CS}
\end{equation}
where $z^{q+}_{i}$ are features in $Q$ that have the same labels of $x$, and $N_{pair}$ is the number of pairs $({F}'(x), z^{Q+}_{i})$.

\subsection{Overall}
Finally, we summarize the teaching loss $\mathcal{L}_{teach}$, the learning loss $\mathcal{L}_{learn}$ and the contrastive similarity loss $\mathcal{L}_{CS}$ to get the final loss $\mathcal{L}_{all}$.
\begin{equation}
    \mathcal{L}_{all} = \mathcal{L}_{teach} + \mathcal{L}_{learn} + \mathcal{L}_{CS}
    \label{all}
\end{equation}
And the whole algorithm is as follows.

\begin{algorithm} 
	\caption{Contrastive Causal Model} 
	\label{alg3} 
	\begin{algorithmic}
		\REQUIRE Teacher $F$ and student $F'$, classifier $C$, knowledge queue $Q$, a batch of data $\mathbb{B}$.
		\STATE Split $\mathbb{B}$ by domain.
		\FOR{$D_{k} \in \mathcal{D}$}
        \STATE $Z \gets F(X^{D_{k}})$
        \STATE Update the parameters of $F'$ with Eq. \ref{update}.
        \STATE $Z' \gets F'(X^{D_{k}})$
        \STATE Teach CCM correlation via $\mathcal{L}_{teach}$ with Eq. \ref{teach}.
        \STATE Increase causal effects by $\mathcal{L}_{learn}$ with Eq. \ref{learn}.
        \STATE Cluster features with the same category by $\mathcal{L}_{CS}$ with Eq. \ref{l_CS}.
        \STATE Calculate $\mathcal{L}_{all}$ with Eq. \ref{all}.
        \STATE Update the parameters of CCM.
        \STATE Pop $z_{m}^{Q}$ and $y_{m}^{Q}$ of $Q$.
        \STATE Push $z'$ and $y'$ into $Q$.
        \ENDFOR
        \RETURN The parameters of CCM.
	\end{algorithmic} 
\end{algorithm}

\section{Experiments}
In this section, we will introduce our experiments in detail. We use $\mathrm{DomainBed}$ \cite{gulrajani2020search} to implement CCM and evaluate performance on three standard datasets: PACS \cite{li2017deeper}, OfficeHome \cite{venkateswara2017deep} and TerraIncognita \cite{beery2018recognition}. And ablation experiments are then performed to demonstrate the lifting effect of each part of the CCM. 

\subsection{Datasets}
PACS \cite{li2017deeper} contains 9,991 images with 4 domains $\{art, cartoon, photo,sketch \}$ and 7 categories.
OfficeHome \cite{venkateswara2017deep} contains 15,588 images with 4 domains $\{art, clipart, product,real\mbox{-}world \}$ and 65 categories.
The settings of TerraIncognita \cite{venkateswara2017deep} remain the same as 
\cite{gulrajani2020search}. It contains 24,788 images with 4 domains $\{L100, L38, L43, L46 \}$ and 10 categories.
\subsection{Implementation Details} 
In our main experiments, we use ResNet50 \cite{he2016deep} as our backbone, and the settings are following $\mathrm{DomainBed}$ \cite{gulrajani2020search}. We use GTX TITAN $\times$ 4 to support the derivation of results. 
Each GTX TITAN graphics card has 12 GB memory. And the cpu is Intel(R) Xeon(R) CPU E5-2660 v3. The version of pytorch is 1.10.0.
In Table \ref{tab:all}, we train CCM with 5 different hyperparameters over 3 times for each test domain. In Table \ref{tab:ablation}, we fix the parameters with the highest accuracy in the validation set. Subsequently, only the loss function $\mathcal{L}_{all}$ of the CCM is adjustable. The size of $Q$ is $4 \times batch\,size \times k$. Both popping and pushing operations are updated in batches. The temperature hyper-parameter $\tau$ is 0.07 and momentum ratio $\alpha$ is 0.999 as same as MoCo.

\subsection{Baselines}
We compare CCM extensively with other DG algorithms. Specially, including Empirical Risk Minimization (ERM) \cite{vapnik1999nature}, Invariant Risk Minimization (IRM) \cite{arjovsky2019invariant}, Group Distributionally Robust Optimization (GroupDRO) \cite{sagawa2019distributionally}, Interdomain Mixup (Mixup)   \cite{yan2020improve}, Marginal Transfer Learning (MTL) \cite{blanchard2017domain}, Meta Learning Domain Generalization (MLDG) \cite{li2018learning}, Maximum Mean Discrepancy (MMD) \cite{li2018domain}, Deep CORAL (CORAL) \cite{sun2016deep}, Domain Adversarial Neural Network (DANN) \cite{ganin2016domain}, Conditional Domain Adversarial Neural Network (CDANN) \cite{li2018deep}, Style Agnostic Networks (SagNet) \cite{nam2021reducing}, Adaptive Risk Minimization (ARM) \cite{zhang2020adaptive}, Variance Risk Extrapolation (VREx) \cite{krueger2021out}, Representation Self-Challenging (RSC) \cite{huang2020self},Smoothed-AND mask (SAND-mask) \cite{shahtalebi2021sand} and  Invariant Gradient Variances for Out-of-distribution Generalization (Fishr) \cite{rame2021fishr}.

\begin{table}[t]
\begin{center}
\adjustbox{max width=\linewidth}{%
\begin{tabular}{lcccc}
\toprule
\textbf{Algorithm}        & \textbf{PACS}             & \textbf{OfficeHome}       & \textbf{TerraIncognita}   & \textbf{Avg}              \\
\midrule
ERM                       & 85.5            & 66.5            & 46.1            & 66.0                      \\
IRM                       & 83.5            & 64.3            & 47.6            & 65.1                      \\
GroupDRO                  & 84.4            & 66.0            & 43.2            & 64.5                      \\
Mixup                     & 84.6            & \underline{68.1}            & \underline{47.9}            & 66.9                      \\
MLDG                      & 84.9            & 66.8            & 47.7            & 66.5                      \\
CORAL                     & 86.2            & 68.7            & 47.6            & 67.5                      \\
MMD                       & 84.6            & 66.3            & 42.2            & 64.4                      \\
DANN                      & 83.6            & 65.9            & 46.7            & 65.4                      \\
CDANN                     & 82.6            & 65.8            & 45.8            & 64.7                      \\
MTL                       & 84.6            & 66.4            & 45.6            & 65.5                      \\
SagNet                    & \underline{86.3}            & \underline{68.1}            & \textbf{48.6}            & \underline{67.7}                      \\
ARM                       & 85.1            & 64.8            & 45.5            & 65.1                      \\
VREx                      & 84.9            & 66.4            & 46.4            & 65.9                      \\
RSC                       & 85.2            & 65.5            & 46.6            & 65.8                      \\
SAND-mask                 & 84.6            & 65.8            & 42.9            & 64.4                      \\
Fishr                     & 85.5            & 67.8            & 47.4            & 66.9                      \\
\midrule
CCM                       & \textbf{87.0}            & \textbf{69.7}            & \textbf{48.6}            & \textbf{68.4}                      \\
\bottomrule
\end{tabular}}
\end{center}
\caption{Results on DomainBed for PACS, OfficeHome and TerraIncognita. The best results are emphasized in bold, and the second best results are marked by underlining.}
\label{tab:all}
\end{table}

\subsection{Result}
We use ResNet50 as backbone to get Table \ref{tab:all} on PACS, OfficeHome and TerraIncognita. It should be noticed that we use default settings of \textrm{DomainBed} and choose the training-domain validation set selection method to filter the final accuracies. We provide more detailed results and visualizations in Appendix.

\textbf{Results on PACS.} The results are shown in Table \ref{tab:all}. CCM not only gets better average accuracy but also achieves the best performance on $\{art, cartoon, sketch\}$ three test domains(see Appendix). In Table \ref{tab:all}, only CCM and SagNet are able to exceed $80\%$ percent accuracy on all test domains. However, the standard deviation of CCM on PACS is less than that of SagNet, but the average accuracy  is greater than that of SagNet. SagNet aims to reduce style bias, just like cutting the causal  path from $D \rightarrow E \rightarrow X$ in Fig \ref{fig:causal-graph} (d). Using our SCM can visually show that $D$ has other paths that affect $X$. And when CCM blocks the causal path from $D$ to $X$ based on our SCM and strengthens the causal effects from $X$ to $Y$, the accuracy is further improved.

\textbf{Results on OfficeHome.}
Table \ref{tab:all} also shows the competitiveness of CCM. On $\{clipart, product,real\mbox{-}world\}$ three test domains, CCM gets the best performance. Compared to other methods, the average accuracy is also the highest among all algorithms and improved by 1.0 points.

\textbf{Results on TerraIncognita.} Although CCM does not achieve the best result  in any individual test domain, it still achieves the best average accuracy on TerraIncognita. This precisely shows that CCM can learn stable causality better and identify it in the target domain in Table \ref{tab:all}.  The highest value does not depend on the performance improvement of a domain alone, but rather on the combined effect of each domain.

In Figure \ref{tab:all}, the results of CCM are better than the other methods on the three datasets. We believe that the controlling of the domain factor $D$ reduces the dependence of the source domain during training. And the causal effect learning helps CCM link unseen images to prior knowledge, making more use of the information in the source domain. The contrastive similarity learning breaks the constraints of the batch. The features belonging to the same category have opportunities to pull in the similarity of each other.

\begin{table}[t]
    \centering
    \adjustbox{max width=\linewidth}{%
    \begin{tabular}{lcccc}
    \toprule
    \textbf{Algorithm}              & \textbf{PACS}        & \textbf{OfficeHome}         & \textbf{TerraIncognita}         & \textbf{Avg}  \\
    \midrule
    ERM                             & 85.5                 & 66.5       & 46.1       & 66.0 \\
    \midrule
    CCM w/o  $\mathcal{L}_{teach}$  & 17.7                 & 1.4        & 14.8      & 11.3 \\
    CCM w/o  $\mathcal{L}_{learn}$  & 86.1                 & 69.7       & 48.4      & 68.1 \\
    CCM w/o  $\mathcal{L}_{CS}$   & 86.5                 & 69.3       & 48.8      & 68.2 \\
    CCM w/o  $\mathcal{L}_{teach}$  w/o  $\mathcal{L}_{learn}$   & 86.2                 & 69.3       & 47.2      & 67.6 \\
    CCM                             & 87.0                 & 69.7       & 48.6      & 68.4 \\
    \bottomrule
    \end{tabular}}
    \caption{Ablation studies of $\mathcal{L}_{teach}$, $\mathcal{L}_{learn}$ and $\mathcal{L}_{CS}$ on PACS, OfficeHome and TerraIncognita.}
    \label{tab:ablation}
\end{table}

\begin{figure*}[ht]
\centering
\includegraphics[width=1\textwidth]{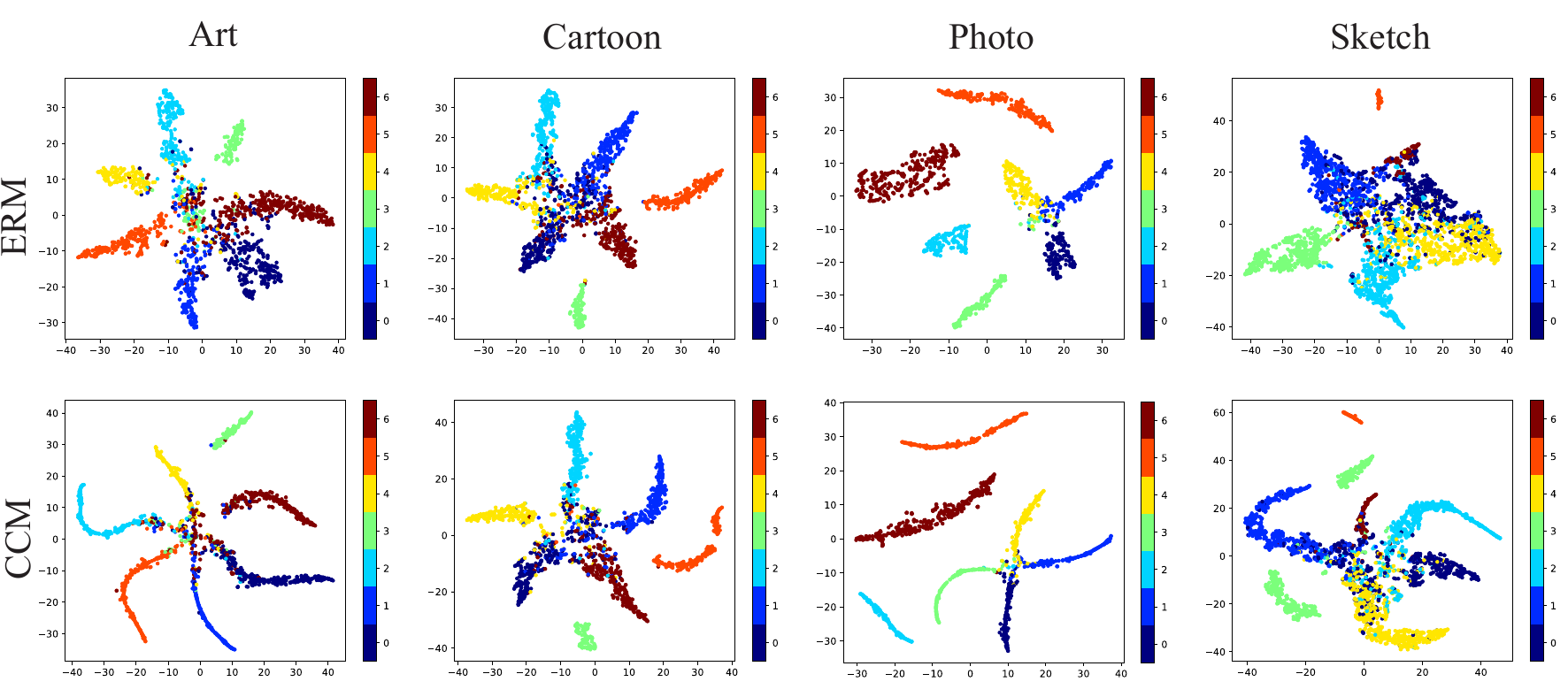}
\caption{The Visualization results of T-SNE \cite{van2008visualizing} on PACS. 
CCM gives clearer boundaries to the categories and makes features belonging to the same category of images more concentrated in feature space $\mathcal{Z}$. Not only the category mixing area of CCM is smaller, but also the distribution of strip shapes shows that CCM prefers to arrange features with the same category in a manifold manner.}
\label{figs:tsne}
\end{figure*}

\subsection{Ablation Study}
We further explore the role of each loss of CCM in the overall training process by ablation study. And the results summarized in Table \ref{tab:ablation}. We can find out that applying $\mathcal{L}_{teach}$ alone can significantly improve the performance of domain generalization, and removing $\mathcal{L}_{teach}$ will cause the accuracies of CCM to drop sharply, close to the accuracy of random predictions. It indicates that CCM learns the basic correlations by $\mathcal{L}_{teach}$ and causal effect learning relies on correlations.

Since causal effect learning requires the assistance of contrastive similarity to complete the quantification of causal effects, $L_{learn}$ and $L_{CS}$ need to exist simultaneously to achieve optimal performance. In Table \ref{tab:ablation}, combining $L_{learn}$ and $L_{CS}$  achieves better results on PACS, OfficeHome, and average. We provide the results of ablation experiments for each domain on PACS, OfficeHome, and TerraIncognita in the Appendix.


\subsection{Visualization}
We use T-SNE \cite{van2008visualizing} to display the classification results of CCM on PACS visually. 
Each column represents a domain and each row represents a model.
To demonstrate how CCM can distinguish different features in space $\mathcal{Z}$, we feed target domain images in CCM to get their corresponding features and apply T-SNE for dimensionality reduction and visualization. The results are presented in Fig \ref{figs:tsne}. We can find CCM clusters features of images that belong to the same category more closely. Compared with ERM, not only the category boundaries of CCM are more explicit, but also the distributions of features that belong to the same category are more concentrated in the feature space $\mathcal{Z}$. Most of the shapes in the CCM visualization results are stripes, meaning CCM learns more accurate similarity relationships between features of the same category and arranges them by a manifold via contrastive similarity learning.


\section{Conclusion}

In this paper, we investigated how to quantify causal effects from images to labels and optimize them. Inspired by the ability of humans to adapt to new environments with prior knowledge, we proposed a novel algorithm called Contrastive Causal Model (CCM), which can transfer unseen images into seen images and quantify the causal effects via the front-door criterion. Through extensive experiments, CCM demonstrates its effectiveness and outperforms current methods. In particular, the prior knowledge of CCM can take many forms, which is the potential to combine with non-image representations to solve multimodal problems.

\bibliography{aaai23}
\end{document}